\documentclass[letterpaper,10pt,conference]{ieeeconf}
\IEEEoverridecommandlockouts
\usepackage{amsmath,amssymb,booktabs,multirow,url}
\usepackage{xcolor}
\usepackage{graphicx}
\usepackage{xspace}
\usepackage{tikz}
\usetikzlibrary{arrows.meta,positioning,calc,fit,backgrounds,shapes.geometric,shapes.misc}
\tikzset{
  frozen/.style={draw=black!55, fill=black!8, rounded corners=2pt, align=center,
    font=\scriptsize, inner sep=3pt, minimum height=6mm},
  trained/.style={draw=orange!85!black, line width=0.7pt, fill=orange!15,
    rounded corners=2pt, align=center, font=\scriptsize, inner sep=3pt, minimum height=6mm},
  cond/.style={draw=blue!55!black, fill=blue!8, rounded corners=2pt, align=center,
    font=\scriptsize, inner sep=3pt, minimum height=5mm},
  latent/.style={draw=black!65, fill=white, rounded corners=3pt, align=center,
    font=\scriptsize, inner sep=2.5pt, minimum height=5mm},
  outp/.style={draw=green!45!black, fill=green!8, rounded corners=2pt, align=center,
    font=\scriptsize, inner sep=3pt, minimum height=6mm},
  geo/.style={draw=black!45, fill=black!4, rounded corners=2pt, densely dotted,
    align=center, font=\scriptsize, inner sep=3pt, minimum height=5mm},
  op/.style={draw=black!60, circle, inner sep=1pt, font=\footnotesize},
  flow/.style={-{Latex[length=2mm]}, draw=black!70, line width=0.5pt},
  swaparr/.style={-{Latex[length=2mm]}, draw=red!70!black, line width=0.9pt},
  panlab/.style={font=\footnotesize\itshape},
}
\newcommand{\legendbox}[1]{\tikz[baseline=-0.5ex]{\node[#1,minimum width=4mm,minimum height=2.6mm,inner sep=0pt]{};}}
\newcommand{\flowsim}{\textsc{FlowSim}\xspace}
\newcommand{\uniad}{\textsc{UniAD}\xspace}
\newcommand{\splatad}{\textsc{SplatAD}\xspace}
\newcommand{\dreamforge}{\textsc{DreamForge}\xspace}
\newcommand{\omnire}{\textsc{OmniRe}\xspace}
\newcommand{\transfuser}{\textsc{TransFuser}\xspace}
\title{FlowSim: Rendering-Free Closed-Loop Simulation for Evaluating End-to-End Driving Systems}
\makeatletter
\begin{document}
\newcommand{\linebreakand}{%
  \end{@IEEEauthorhalign}
  \hfill\mbox{}\par
  \mbox{}\hfill\begin{@IEEEauthorhalign}
}
\makeatother
\author{
\authorblockN{1\textsuperscript{st} Hanfeng Wu}
\authorblockA{\textit{Karlsruhe Institute of Technology} \\
\textit{BMW Group}\\
Munich, Germany \\
hanfeng.wu@bmw.de}
\and
\authorblockN{2\textsuperscript{rd} Alvaro Marcos-Ramiro}
\authorblockA{\textit{BMW Group} \\
Munich, Germany \\
alvaro.marcos-ramiro@bmw.de}
\and
\authorblockN{3\textsuperscript{th} Michael Schmidt}
\authorblockA{\textit{BMW Group} \\
Munich, Germany \\
michael.se.schmidt@bmw.de}
\linebreakand
\authorblockN{4\textsuperscript{th} Christoph Stiller}
\authorblockA{\textit{Karlsruhe Institute of Technology} \\
Karlsruhe, Germany \\
stiller@kit.edu}
}
\maketitle

\thispagestyle{empty}
\pagestyle{empty}
\begin{abstract} Closed-loop evaluation of end-to-end (E2E) driving systems typically relies on generating sensor observations and re-processing them through the perception encoder, introducing substantial computational cost and a sim2real gap between simulated and real sensor observations. In this paper, we ask a more fundamental question: is image-space simulation necessary for closed-loop evaluation of E2E driving systems? We propose \textsc{FlowSim}, a rendering-free simulation paradigm that predicts the internal latent representation directly from a previous latent and target-frame symbolic scene conditions. The predicted latent is then processed by the original frozen downstream modules, eliminating image rendering and perception re-encoding. Explicit cross-frame agent association improves latent transition prediction and rollout stability, while variable-horizon simulation enables long-horizon prediction from a common real anchor without accumulating intermediate rollout errors. 
Experiments on nuScenes and NavSim demonstrate that \textsc{FlowSim} preserves downstream behavior competitively with, and often exceeds image-space simulation in source-to-reference agreement across variable simulation horizons and both dense and sparse latent representations.
By predicting the latent directly, both instantiations of \textsc{FlowSim} simulate each step in 0.04--0.06s, substantially faster than the image-space simulation methods they are compared against.
These results establish \flowsim as an efficient and practical alternative to conventional image-space simulation methods. Our project page is \url{https://hanfeng-wu.github.io/flowsim_page/}.
\end{abstract}

\section{INTRODUCTION}

Simulation plays a central role in the development and validation of E2E driving systems. Existing evaluation frameworks~\cite{navsim,pseudosim,nuscenes} can assess driving behavior under evolving scene conditions, but can not update the internal state of the driving system during rollout. In contrast, sensor simulation approaches generate future sensor observations and re-process them through the perception stack, allowing the internal state of the system to evolve over time. As illustrated in Fig.~\ref{fig:compare}, both neural reconstruction approaches and driving world models ultimately rely on an observation-space simulation pipeline before decisions can be produced. However, E2E driving systems do not make decisions directly from pixels. Instead, observations are transformed into an intermediate latent representation from which downstream tasks are derived. This raises a fundamental question: \begin{quote} \emph{Is image-space simulation actually necessary for closed-loop evaluation of E2E driving systems?} \end{quote}
\begin{figure}[t]
\centering
\resizebox{\columnwidth}{!}{%
\begin{tikzpicture}[font=\footnotesize, >={Latex[length=1.8mm]},
    node distance=3mm, every node/.style={inner sep=2.6pt}]
  \newcommand{\condtext}{\textbf{Scene}\\\textbf{conditions}\\[1pt]
     $\bullet$ agents states\\
     $\bullet$ HD map\,\textit{(opt.)}\\
     $\bullet$ ego states}
  \newcommand{\headstext}{\textbf{Frozen}\\\textbf{heads}\\[3pt]{\scriptsize det\,/\,seg\\ mot\,/\,plan}}

  \node[cond, align=left, text width=1.7cm] (cond1) at (0,0) {\condtext};
  \node[trained] (rend) at (3.55,0)  {Neural\\renderer};
  \node[latent]  (img)  at (5.2,0)   {rendered\\image};
  \node[frozen]  (enc)  at (6.85,0)  {Frozen\\encoder};
  \node[frozen, align=center, text width=1.55cm] (heads1) at (8.75,0) {\headstext};
  \node[font=\footnotesize\bfseries, anchor=south west] at (2.5,0.55)
    {(i) Neural reconstruction};
  \draw[flow, blue!55!black] (cond1.east) -- (rend.west);
  \draw[flow] (rend) -- (img);
  \draw[flow] (img)  -- (enc);
  \draw[flow] (enc.east) -- node[above, font=\scriptsize] {$\hat Z_t$} (heads1.west);

  \node[cond, align=left, text width=1.7cm] (cond2) at (0,-2.2) {\condtext};
  \node[latent]  (pimg) at (1.85,-2.2)  {prev.\\image $I_{t-1}$};
  \node[trained] (wm)   at (3.5,-2.2)   {World\\model};
  \node[latent]  (dimg) at (5.2,-2.2)   {predicted\\image};
  \node[frozen]  (enc2) at (6.85,-2.2)  {Frozen\\encoder};
  \node[frozen, align=center, text width=1.55cm] (heads2) at (8.75,-2.2) {\headstext};
  \node[font=\footnotesize\bfseries, anchor=south west] at (2.5,-1.65)
    {(ii) World model};
  \draw[flow, blue!55!black] ([yshift=-7mm]cond2.east) -| (wm.south);
  \draw[flow] (pimg) -- (wm);
  \draw[flow] (wm)   -- (dimg);
  \draw[flow] (dimg) -- (enc2);
  \draw[flow] (enc2.east) -- node[above, font=\scriptsize] {$\hat Z_t$} (heads2.west);

  \node[cond, align=left, text width=1.7cm] (cond3) at (0,-4.4) {\condtext};
  \node[latent]  (plat) at (1.85,-4.4)  {prev.\\latent $Z_{t-1}$};
  \node[trained] (fs)   at (4.2,-4.4)   {\flowsim};
  \node[frozen, align=center, text width=1.55cm] (heads3) at (8.75,-4.4) {\headstext};
  \node[font=\footnotesize\bfseries, anchor=south west] at (2.5,-3.85)
    {(iii) \flowsim\ \textnormal{(ours)}};
  \draw[flow, blue!55!black] ([yshift=-7mm]cond3.east) -| (fs.south);
  \draw[flow] (plat) -- (fs);
  \draw[flow] (fs.east) -- node[above, font=\scriptsize] {$\hat Z_t$ \textnormal{(direct)}}
      node[below, font=\scriptsize\itshape, text=black!55] {no rendering}
      (heads3.west);

  \draw[black!35, densely dashed, line width=0.4pt] (-1.05,-1.1) -- (9.65,-1.1);
  \draw[black!35, densely dashed, line width=0.4pt] (-1.05,-3.3) -- (9.65,-3.3);
\end{tikzpicture}%
}
\vspace{1mm}
{\scriptsize
\legendbox{cond}~scene conditions\quad
\legendbox{trained}~source generator\quad
\legendbox{frozen}~frozen e2e stack\quad
}

\caption{Comparison of our simulation paradigm (\flowsim) with 2 existing closed-loop simulation protocols for evaluating E2E driving systems, each producing a latent $\hat Z_t$ that is sourced by the \emph{same} frozen downstream
heads. \textbf{(i)}~\emph{Neural reconstruction} renders the target frame from the explicit scene
conditions (agent boxes, ego pose) into pixels, which the frozen encoder
re-encodes into $\hat Z_t$. \textbf{(ii)}~A \emph{world model} likewise produces pixels for the
\emph{same} encoder, conditioned on its own \emph{previous images} $I_{t-1}$ together with the scene
conditions. \textbf{(iii)}~\flowsim\ is conditioned on the
\emph{previous latent} $Z_{t-1}$ and the same scene conditions, yet predicts the latent
$\hat Z_t$ \emph{directly}, bypassing image rendering and re-encoding.}
\label{fig:compare}
\end{figure}
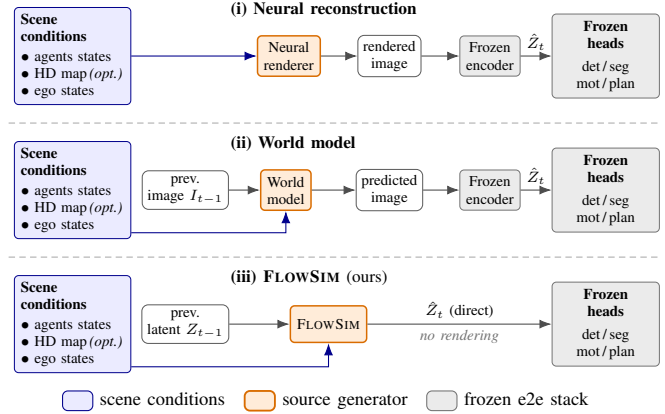

In this paper, we argue that it is not. We introduce \flowsim, a rendering-free simulation paradigm that evaluates an E2E driving system directly through its internal latent representation. Instead of generating images and re-running the perception encoder, \flowsim predicts the latent state expected by the downstream modules from previous frame latent and target frame symbolic scene conditions, including ego state, map context, and surrounding agents. The predicted latent is then processed by the original frozen downstream stack without modification. This formulation replaces the conventional closed-loop simulation pipeline \[ \text{Scene} \rightarrow \text{Rendering} \rightarrow \text{Perception} \rightarrow \text{Planning}, \] with \[ \text{Scene} \rightarrow \text{Latent Simulation} \rightarrow \text{Planning}. \] The goal therefore shifts from reconstructing observations to preserving the behavior of the target driving system.

Realizing this paradigm requires the simulator to model how the scene changes between the source and target frames. Target-frame agent states describe the targeted scene but do not explicitly encode which agents persist across time. \textsc{FlowSim} therefore introduces associated-agent conditioning, which augments each target agent with its source-frame correspondence and relative motion. This converts temporal correspondence from an implicit inference problem into an explicit conditioning signal.
The same conditional formulation also supports simulation across different temporal horizons. Conventional autoregressive rollout feeds each generated latent back into the next transition, mixing future-state prediction with compounding errors under generated-to-generated inputs. \textsc{FlowSim} instead supports real-anchored variable-horizon simulation: a future latent at a selected horizon is predicted directly from a logged latent anchor and the symbolic scene conditions at the target time. Variable-horizon training allows a single simulation step to handle these different horizons while preserving the real-to-simulated transition used during training.

\textsc{FlowSim} is intended as a simulation paradigm rather than a specific generative architecture: we instantiate it on both the dense bird's-eye-view (BEV) latent used by UniAD~\cite{uniad} and the sparse latent representation used by TransFuser~\cite{transfuser} (Sec.~\ref{sec:formulation}), and validate on nuScenes and NavSim that it preserves downstream behavior competitively with, and often more faithfully than, image-space simulation baselines while simulating each step in only 0.04--0.06s.

Our contributions are summarized as follows: 
\begin{itemize} 
\item We introduce \textsc{FlowSim}, a rendering-free latent-space simulation paradigm for closed-loop evaluation of E2E driving systems, eliminating image generation while achieving simulation speeds of 0.04--0.06s per step.
\item We propose agent association conditioning, which explicitly encodes cross-frame agent correspondence and relative motion for latent transition prediction.
\item We introduce variable-horizon simulation, supported by variable-horizon training, which predicts future latent states directly from a common real anchor while avoiding drift through intermediate generated latents.
\item We establish a source-to-reference agreement protocol and validate the complete framework across dense UniAD and sparse TransFuser latent representations on nuScenes and NavSim.
\end{itemize}

\section{RELATED WORK}
\subsection{Neural Reconstruction of Driving Scenes}
Recent work on Neural fields and 3D Gaussian Splatting~\cite{gaussian_splats} has been used to reconstruct dynamic urban scenes by decomposing static background from moving actors, for sensor simulation~\cite{street_gaussians,omnire,neurad,splatad,nfl,dynfl}. We use two such reconstructions, \omnire~\cite{omnire} and \splatad~\cite{splatad}, as image sources in our own evaluation. These methods deliver high visual fidelity, but they operate in raw sensor space: any E2E stack under evaluation must re-encode the rendered frames through its full perception backbone. As a result, simulation fidelity depends on both the quality of the rendered observations and how faithfully they produce the same latent representation as real sensor data. They are also fit per scene to a logged trajectory, so when the ego deviates from the original log, the simulator must extrapolate unseen viewpoints beyond the recorded data.

\subsection{World Models and Latent Simulation for Driving} Recent driving world models represent future scenes in various forms, including multiview video observations \cite{drivewm,dreamforge,panacea,magicdrive_v2,cosmos_drive,gaia2,nwm}, dense or sparse latent representations~\cite{bevworld,sparseworld}, occupancy states~\cite{occworld}, and planner-oriented latent features \cite{law,world4drive,autojepa}. Despite their differences in representation, most methods learn the scene representation and forecasting model jointly, treating the representation itself as part of the world model. By contrast, \flowsim does not define its own predictive representation. Instead, it directly simulates the latent representation produced by an independently trained and frozen driving stack.

Temporal consistency of dynamic agents presents another challenge. Most controllable driving generators represent agents using frame-wise symbolic states and leave temporal correspondence across frames to be inferred implicitly by the model~\cite{panacea,gaia2,cosmos_drive}. TrackDiffusion \cite{trackdiffusion} introduces explicit instance-level conditioning through image-space tracklets to improve appearance consistency across generated videos. \flowsim instead addresses this by explicitly conditioning on cross-frame agent correspondence and relative motion~(Sec.~\ref{sec:assoc}).

Long-horizon prediction remains another central challenge. Existing approaches typically either improve autoregressive stability through rollout-aware training and self-correction mechanisms, as explored by HorizonDrive~\cite{horizondrive}, or avoid recursive rollout by predicting multiple future steps jointly, as in Drive-WM~\cite{drivewm}, DWM~\cite{dwm} and DreamForge~\cite{dreamforge}. \flowsim instead supports real-anchored variable-horizon simulation across multiple prediction horizons using only single prediction step~(Sec.~\ref{sec:timegap}).

\section{METHOD}
\begin{figure*}[t]
    \centering
    \includegraphics[width=1\linewidth]{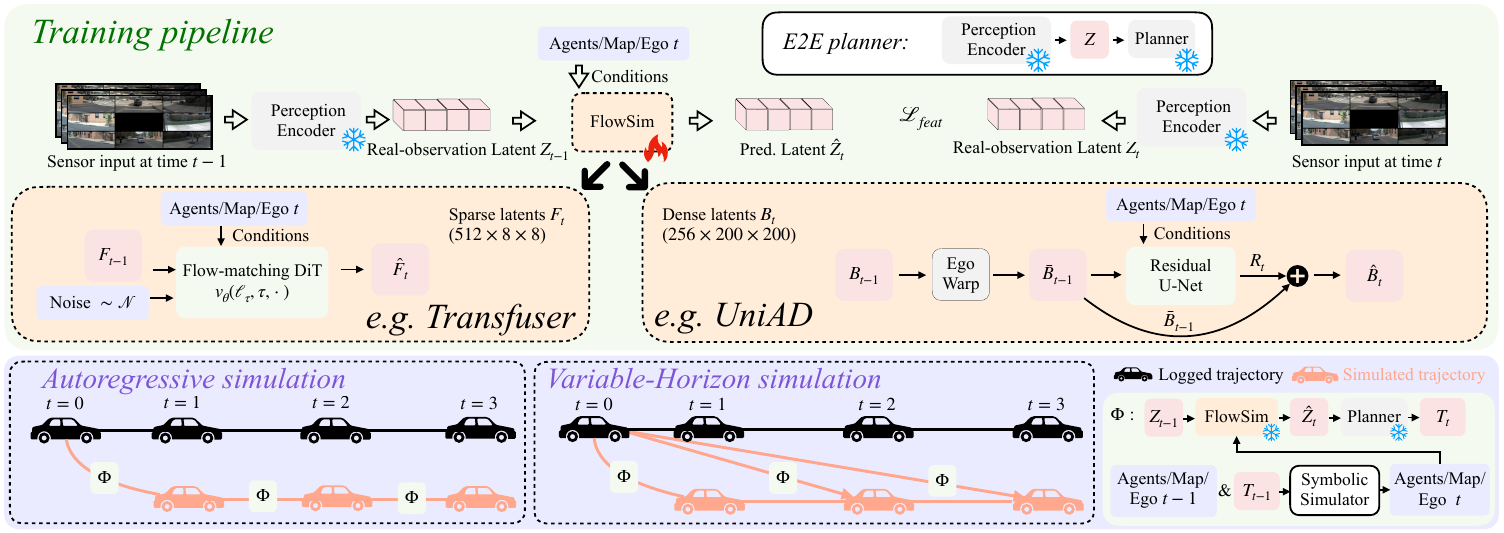}
    \vspace{-6mm}
    \caption{Overview of \textsc{FlowSim}. \textbf{Top: Training.} A frozen perception encoder extracts the source and target latents, $Z_{t-1}$ and $Z_t$, from real sensor observations. Given $Z_{t-1}$ and target-frame symbolic scene conditions, \textsc{FlowSim} predicts $\hat{Z}_t$ and is trained to reconstruct the target latent $Z_t$. The same formulation is instantiated with a residual U-Net for UniAD's dense BEV representation and a conditional flow-matching model for TransFuser's sparse latent representation. \textbf{Bottom: Simulation protocols.} \flowsim supports both autoregressive simulation and variable-horizon simulation. In the latter, multiple future horizons are simulated directly from a common real latent anchor, preserving the real-to-simulated transition used during training and avoiding intermediate autoregressive drift. The predicted latent is passed directly to the frozen downstream stack to obtain the ego trajectory. Both simulation protocols can be coupled with a symbolic scene simulator that updates the ego and surrounding agents from the planned trajectory, enabling closed-loop simulation without image rendering.}
    \label{fig:main}
    \vspace{-3mm}
\end{figure*}

\subsection{General Latent-Simulation Formulation}\label{sec:formulation}
\flowsim formulates simulation as direct latent prediction inside a frozen E2E driving stack. Rather than generating sensor observations and re-running the perception encoder, \flowsim predicts the target latent directly from a previous latent and symbolic target-scene conditions. Let $Z_t$ denote the internal latent representation produced by a frozen E2E driving system at time $t$, immediately before its downstream heads. Depending on the target architecture, $Z_t$ may take the form of a dense bird's-eye-view (BEV) representation, as in UniAD, or a sparse latent feature set, as in TransFuser. Given the latent state from the previous timestep and the target scene conditions, \flowsim predicts the next latent according to 
\begin{equation} 
\hat{Z}_t = f_{\theta}(Z_{t-1}, A_t, M_t, e_t), \label{eq:flowsim} 
\end{equation} 
where $A_t$ denotes surrounding agent condition, $M_t$ denotes map context, and $e_t$ denotes ego-state change. The simulator $f_{\theta}$ is the only trainable component. All downstream modules remain frozen throughout training and evaluation.

As illustrated in Fig.~\ref{fig:main}, the simulator is trained using latent representations extracted from the target driving stack. During training, the frozen perception encoder generates paired latent states $(Z_{t-1}, Z_t)$ from real sensor observations. \flowsim learns to predict $Z_t$ from $Z_{t-1}$ and the target-frame scene conditions. During evaluation, only $Z_{t-1}$ and the scene conditions are available; the predicted latent $\hat{Z}_t$ is injected directly into the frozen downstream stack.

A key advantage of this formulation is that it is agnostic to the structure of the latent representation: the simulator interface is identical whether the target latent is dense or sparse, decoupling the simulation paradigm from the choice of generative architecture. In this work, we instantiate $f_{\theta}$ using a residual U-Net for dense BEV representations and a conditional flow-matching~\cite{lipman_flow_matching} DiT for sparse latent representations, while keeping the conditioning, training, and evaluation framework unchanged. The rest of this section details the condition of cross-frame agent correspondence~(Sec.~\ref{sec:assoc}), two concrete instantiations of $f_\theta$ for dense and sparse latents~(Secs.~\ref{sec:dense} and~\ref{sec:sparse}), the shared training objective~(Sec.~\ref{sec:loss}) and variable-horizon training and simulation protocol~(Sec.~\ref{sec:timegap}).

\subsection{Associated Dynamic-Agent Conditioning}\label{sec:assoc}
\begin{figure}
    \centering
    \includegraphics[width=1\linewidth]{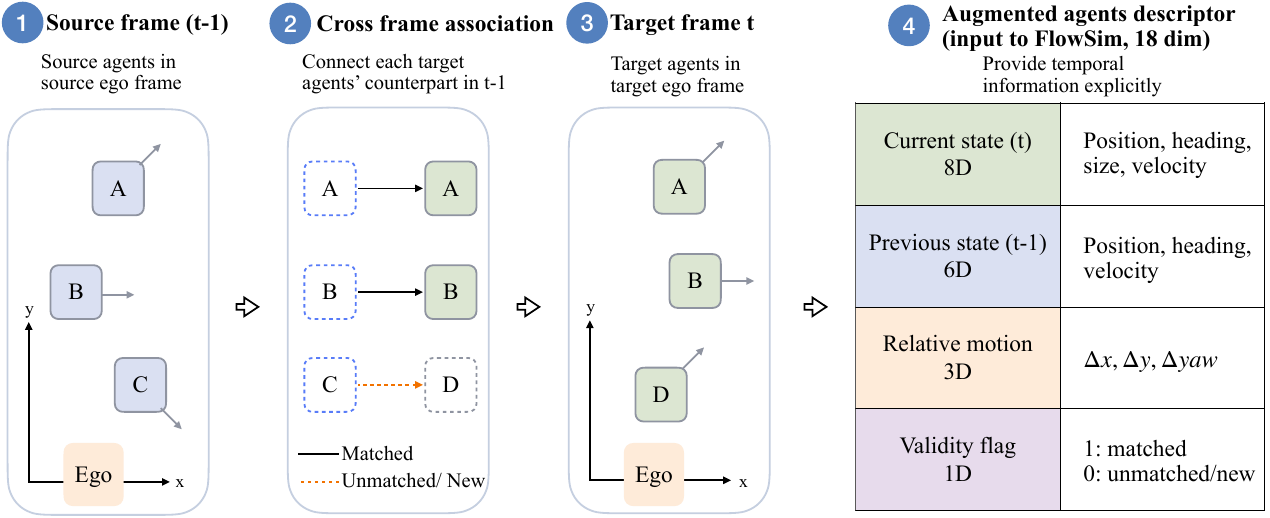}
    \vspace{-6mm}
    \caption{Associated dynamic-agent conditioning. Each target-frame agent is augmented with information from its corresponding source-frame agent. The resulting descriptor combines the target state, source state, relative motion, and a validity flag, explicitly exposing cross-frame correspondence and temporal dynamics to the latent simulator. New agents without a valid correspondence are retained with an invalid flag rather than removed.}
    \label{fig:assoc}
    \vspace{-3mm}
\end{figure}

Target-frame agent states describe the target scene but do not explicitly encode cross-frame agent correspondence. Most controllable driving world models similarly represent dynamic agents as independent frame-wise symbolic states, leaving temporal correspondence to be inferred implicitly by the model. As a result, the simulator must determine which source-frame agent each target-frame agent corresponds to, while simultaneously predicting the latent transition. This ambiguity becomes increasingly important as the source-to-target interval grows. 

We address this issue by converting temporal correspondence from an implicit inference problem into an explicit conditioning signal. Instead of providing independent per-frame agent states, we augment each target-frame agent with information from its corresponding source-frame agent and the motion connecting the two states (Fig.~\ref{fig:assoc}). For each target-frame agent with state
\begin{equation}
s_i^t =
[x_i,y_i,yaw_i,
size_i,velocity_i],
\end{equation}
we construct an associated descriptor $a_i^t \in A_t$, with
\begin{equation}
a_i^t
=
[s_i^t,\;
s_i^{t-1},\;
\Delta s_i,\;
c_i],
\end{equation}
where $s_i^{t-1}$ denotes the corresponding source-frame agent,
$\Delta s_i$ describes their relative motion, and $c_i$
indicates whether a valid correspondence exists. 

Such agent association descriptor provides explicit exposure of cross-frame correspondence and motion history to the latent simulator. The simulator no longer needs to recover temporal relationships implicitly from the scene layout and can instead focus on modeling the latent transition. This conditioning formulation is shared by both dense and sparse \textsc{FlowSim} instantiations and improves both single-step and autoregressive rollout stability.

\vspace{-1mm}
\subsection{Sparse Instantiation: Conditional Flow Matching} \label{sec:sparse} We first instantiate \textsc{FlowSim} on sparse latent representations such as the fused token representation used by \transfuser. Given the previous latent $F_{t-1}$ and the target-frame scene conditions, the simulator predicts the target latent through a DiT-based velocity field trained with conditional flow matching (Fig.~\ref{fig:main}). Specifically, \begin{equation} v_\theta(\ell_\tau,\tau,F_{t-1},A_t,M_t,e_t) \end{equation} is trained to transport a noise sample to the target latent $F_t$ along a probability path indexed by $\tau\in[0,1]$. The predicted latent $\hat F_t$ is obtained by integrating the learned flow field at inference time. Following FlowDrive~\cite{flowdrive}, agent, map, and ego states are each encoded to token sequences via lightweight MLPs/CNN and fused via cross-attention. Flow-matching DiT provides an effective generative framework for token-based latent representations and serves here as one instantiation of the general simulator $f_\theta$.
\vspace{-1mm}
\subsection{Dense Instantiation: Residual BEV Prediction} \label{sec:dense} Dense BEV representations such as those used by \uniad pose a different challenge. Unlike sparse latent features, BEV features contain strong spatial structure, and an ego warp typically provides a strong approximation of the target latent. In this setting, the primary objective is not generating a new latent from scratch, but predicting localized scene changes conditioned on agent motion. We therefore instantiate $f_\theta$ as an ego-motion-aligned residual predictor (Fig.~\ref{fig:main}). The previous BEV latent $B_{t-1}$ is first warped into the target ego frame using the relative ego motion in $e_t$, producing an aligned feature map $\bar B_{t-1}$. The residual predictor then estimates the residual latent change $R_t$ with  \begin{equation} \hat B_t = \bar B_{t-1} + R_t . \label{eq:residual} \end{equation} Residual prediction formulates the task as learning deviations from an ego warp, focusing model capacity on dynamic content rather than reconstructing the entire BEV representation. We implement the residual predictor using a standard Residual U-Net. The U-Net downsamples $\bar B_{t-1}$ through four resolution stages ($200\!\to\!100\!\to\!50\!\to\!25$), injecting agent, map, and status tokens via cross-attention at the $25\times25$ bottleneck with spatial tokens.

\vspace{-1mm}
\subsection{Training Objective} \label{sec:loss} \flowsim is trained to predict the latent representation produced by a frozen E2E driving stack. Since the frozen downstream stack operates directly on the latent representation, training \flowsim to reconstruct the target latent provides a direct objective without requiring additional task-specific supervision. We therefore optimize a simple latent reconstruction objective, \begin{equation} \mathcal{L} = \|\hat Z_t-Z_t\|_1. \end{equation}The same objective is used for both dense and sparse instantiations. All downstream modules remain frozen throughout training and are used only during evaluation.
\vspace{-1mm}
\subsection{Variable-Horizon Training and Simulation} \label{sec:timegap} Autoregressive rollout predicts a long horizon through a sequence
of generated-to-generated transitions. In contrast,
variable-horizon simulation predicts a target latent directly from a
latent extracted from a logged frame. The source and target
frames need not be adjacent, allowing the same anchor to support
predictions at different future horizons.

To train a single simulator for these variable-horizon
transitions, 
we constructed training pairs using multiple source-target intervals $\Delta t$ from the supported horizon range, and the source latent
$Z_{t-\Delta t}$, target latent $Z_t$, associated-agent
descriptors $A_t^{(\Delta t)}$, map context $M_t$, and accumulated
ego-state change $e_t^{(\Delta t)}$ are constructed for that
interval. The training objective becomes
\begin{equation}
\mathcal{L}_{\text{VH}} = \left\| f_\theta \left( Z_{t-\Delta t}, A_t^{(\Delta t)}, M_t, e_t^{(\Delta t)} \right) - Z_t \right\|_1 .
\label{eq:variable_gap}
\end{equation}
Sampling $\Delta t$ exposes the simulator to different amounts of
ego motion, agent motion, and scene change within a single
training run. At evaluation time, the trained model can directly
predict a target latent at any supported horizon from the nearest
available real anchor, without recursively generating the
intermediate latent states. This keeps each transition consistent
with the real-to-simulated input pattern used during training.
Because each target horizon is predicted independently, the conditioning inputs can be updated online as the simulated scene evolves. This makes variable-horizon simulation naturally compatible with closed-loop deployment, unlike clip-based video generators~\cite{drivewm,dwm,dreamforge} that pre-generate a fixed future sequence and cannot easily incorporate intermediate updates of evolving scenes.

\section{EVALUATION PROTOCOL}\label{sec:eval}
\subsection{Baselines}\label{sec:baselines}
We evaluate dense latent simulation on nuScenes with a frozen \uniad stack and sparse latent simulation on NavSim/OpenScene with a frozen \transfuser stack. On nuScenes, we compare \flowsim with two image-space sources: \splatad~\cite{splatad}, a per-scene neural reconstruction method, and \dreamforge~\cite{dreamforge}, a conditional driving video model. We also include a training-free \emph{warp} baseline that ego-aligns the previous real BEV but predicts no residual ($R_t{=}0$ in Eq.~\eqref{eq:residual}). On NavSim, we compare \flowsim with \omnire~\cite{omnire}, a neural reconstruction method for urban scenes. All simulation sources have access to the same target-frame symbolic conditions: ego, map, agent.
The comparison therefore isolates the simulation source itself rather than differences in the available scene information.
The latents generated by all sources are evaluated through the same frozen stack for each dataset, using only frame samples valid for every source in the corresponding comparison.
\vspace{-1mm}
\subsection{Source-to-Reference Agreement}\label{sec:agreement}
We evaluate simulation sources by how faithfully they preserve the behavior of a frozen driving stack, rather than by image realism, latent reconstruction accuracy, or task accuracy against ground-truth annotations. The reference branch encodes real observations to a latent $Z_t$ and then the frozen heads yield the results of downstream tasks. The source branch differs only in how the latent is obtained: an image-space source re-encodes generated observations through the same encoder, whereas \flowsim injects its predicted $\hat Z_t$ directly. Different latent sources are decoded by the same frozen heads and compared through task-specific agreement metrics.

Let $P_h$ be a frozen head, $h$ indexing detection, segmentation, motion, or planning. We define source-to-reference agreement as
\begin{equation}
D_h = d_h\!\left(P_h(\hat Z_t),\,P_h(Z_t)\right),
\label{eq:agreement}
\end{equation}
with $d_h$ a task-specific discrepancy. This measures how replacing the simulation source changes the frozen stack's behavior, not accuracy against annotations: the stack may itself err, and an ideal simulation source behaves as if the real latent were present.

For planning we compare trajectories over the stack's own planner horizon (\uniad: six waypoints, $3.0$s; \transfuser: eight, $4.0$s). With $N$ waypoints,
\begin{equation}
\mathrm{ADE}=\frac{1}{N}\sum_{k=1}^{N}\|\hat p_k-p_k\|_2,\qquad
\mathrm{FDE}=\|\hat p_N-p_N\|_2,
\end{equation}
where $\hat p_k$ and $p_k$ produced by simulated and real latents through the same planner; these are agreement errors, not errors against a GT trajectory. We additionally report, for \uniad: detection IoU and center ADE, lane and drivable-area IoU, and motion prediction and planning ADE/FDE; for \transfuser: BEV segmentation pixel and mIoU agreement, detection box-center L2 and heading error, and the same planning ADE/FDE.

\vspace{-1mm}
\subsection{Simulation Protocols}\label{sec:protocols}
\paragraph{\flowsim.}
We evaluate the two protocols introduced in Sec.~\ref{sec:timegap}: variable-horizon (VH) simulation, which predicts $\hat Z_t$ directly from a real anchor $Z_{t-\Delta t}$ (Eq.~\eqref{eq:variable_gap}) and can thus reach multiple horizons from a common real anchor without recursively consuming generated latents, and autoregressive (AR) simulation, which instead feeds each predicted latent back as the next source, introducing a sim-to-sim shift absent from training.
\paragraph{Image-space sources on nuScenes.}
The image-space baselines require source-specific evaluation protocols because they differ substantially in how future frames and temporal context are generated. \splatad reconstructs a logged scene and does not extrapolate in time, so $\Delta t$ is not applicable. We evaluate \splatad~\cite{splatad} in two settings: \emph{gt history}, where real images build \uniad's history and only the target images are rendered, and \emph{generated history}, where rendered images are used for history BEV. \dreamforge~\cite{dreamforge} supports temporal extrapolation by jointly generating a future video clip rather than recursively feeding back individual frames. For horizons beyond a single clip~($>$1.33s), rollouts are constructed by chaining generated clips, using the last frame of one clip to initialize the next. The resulting frames are processed by the frozen \uniad encoder and downstream modules. At $\Delta t=0.5$s, both \dreamforge and \flowsim are conditioned on the same logged history. However, \dreamforge benefits from \uniad's native image encoder when constructing the BEV features, whereas \flowsim must predict the target BEV directly. At longer horizons, \dreamforge operates on its own generated history, introducing accumulated image-generation and re-encoding errors.

\paragraph{Image-space source on NavSim.}
\transfuser is a single-frame planner with no streaming latent history. We therefore evaluate \omnire~\cite{omnire} by rendering each target frame independently, passing it through the frozen \transfuser encoder and heads, and comparing the outputs with those computed by the corresponding real frame over all common valid frames.
\begin{table*}[t]
\caption{
Multi-horizon downstream agreement with UniAD under the protocol in Sec.~\ref{sec:protocols}. FlowSim uses real-anchored latent prediction; DreamForge and SplatAD are image-space sources, and warp is a training-free ego-motion baseline. SplatAD does not extrapolate temporally, so its gt-history/generated-history results are repeated for reference. $\dagger$ indicates the CPU offload, because of insufficient VRAM on A100(40GB). Bold indicates the best result at each horizon.
}
\vspace{-1mm}
\label{tab:agreement_gtprev}
\centering\footnotesize
\setlength{\tabcolsep}{3.2pt}
\begin{tabular}{llcccccccccc}
\toprule
& & \multicolumn{2}{c}{Detection} & \multicolumn{2}{c}{Map} & \multicolumn{2}{c}{Motion} & \multicolumn{2}{c}{Planning} & \multicolumn{2}{c}{Inference} \\
\cmidrule(lr){3-4}\cmidrule(lr){5-6}\cmidrule(lr){7-8}\cmidrule(lr){9-10}\cmidrule(lr){11-12}
Source & $\Delta t$ & IoU$\uparrow$ & cADE$\downarrow$ & Lane$\uparrow$ & Drive.$\uparrow$ & ADE$\downarrow$ & FDE$\downarrow$ & ADE$\downarrow$ & FDE$\downarrow$ & Time\,(s)$\downarrow$ & VRAM\,(GB)$\downarrow$ \\
\midrule
\splatad (gt hist.)~\cite{splatad} & \multirow{4}{*}{$0.5$~s} & \textbf{0.75} & 1.86 & \textbf{0.48} & 0.84 & 6.03 & 11.03 & 0.31 & 0.64 & 0.13 & 4.5\\
\dreamforge (gt hist.)~\cite{dreamforge}  & & 0.65 & 2.32 & \textbf{0.48} & 0.85 & 6.35 & 11.57 & \textbf{0.30} & \textbf{0.61} & 92.3  & 38.2\\
Warp ($R_t{=}0$) & & 0.57 & 1.89 & 0.46 & 0.82 & \textbf{5.87} & \textbf{10.72} & 0.62 & 1.14 &  \textless\,0.01& -\\
\flowsim (VH, assoc.) & & 0.68 & \textbf{1.69} & \textbf{0.48} & \textbf{0.86} & 5.96 & 10.88 & 0.53 & 0.97 & 0.04 & 3.0\\
\midrule
\splatad (gen.\ hist.) & \multirow{4}{*}{$1.5$~s} & 0.55 & 4.65 & 0.35 & 0.73 & 8.13 & 14.87 & \textbf{0.61} & 1.19 & 0.13 & 4.5\\
\dreamforge (gen.\ hist.) & & 0.38 & 6.17 & 0.27 & 0.69 & 8.92 & 16.29 & 0.99 & 2.08 & 267.8 &40.0\\
Warp ($R_t{=}0$) & & 0.37 & 3.94 & 0.34 & 0.69 & 7.70 & 13.95 & 1.35 & 2.41 & \textless\,0.01 & - \\
\flowsim (VH, assoc.) & & \textbf{0.58} & \textbf{1.99} & \textbf{0.43} & \textbf{0.82} & \textbf{7.09} & \textbf{12.86} & \textbf{0.61} & \textbf{1.14} & 0.04 & 3.0\\
\midrule
\splatad (gen.\ hist.) & \multirow{4}{*}{$2.5$~s} & 0.55 & 4.65 & 0.35 & 0.73 & 8.13 & 14.87 & 0.61 & 1.19 & 0.13 & 4.5\\
\dreamforge (gen.\ hist.) & & 0.38 & 5.91 & 0.26 & 0.67 & 8.99 & 16.41 & 0.97 & 2.03 & 401.5 & 25.8$\dagger$\\
Warp ($R_t{=}0$) & & 0.31 & 5.33 & 0.27 & 0.61 & 8.04 & 14.53 & 2.03 & 3.61 & \textless\,0.01 & -\\
\flowsim (VH, assoc.) & & \textbf{0.56} & \textbf{2.12} & \textbf{0.40} & \textbf{0.81} & \textbf{7.24} & \textbf{13.14} & \textbf{0.60} & \textbf{1.13} & 0.04 & 3.0\\
\midrule
\splatad (gen.\ hist.) & \multirow{4}{*}{$4.0$~s} & \textbf{0.55} & 4.65 & 0.35 & 0.73 & 8.13 & 14.87 & \textbf{0.61} & \textbf{1.19} & 0.13 & 4.5\\
\dreamforge (gen.\ hist.) & & 0.36 & 6.16 & 0.24 & 0.65 & 8.83 & 16.11 & 0.99 & 2.07 & 726.0 & 25.8$\dagger$\\
Warp ($R_t{=}0$) & & 0.24 & 6.81 & 0.21 & 0.52 & 8.17 & 14.75 & 2.77 & 4.88 & \textless\,0.01 & -\\
\flowsim (VH, assoc.) & & \textbf{0.55} & \textbf{2.18} & \textbf{0.38} & \textbf{0.79} & \textbf{7.44} & \textbf{13.50} & 0.66 & 1.23 & 0.04 & 3.0\\
\bottomrule
\end{tabular}
\end{table*}
\begin{table}[t]
\caption{Downstream agreement with \transfuser on NavSim \texttt{navtest}. We compare the image-space \omnire reconstruction with \flowsim under autoregressive (AR) and real-anchored variable-horizon (VH) simulation, with and without agent association. Each step corresponds to a $0.5$~s keyframe ($n{=}4$ denotes $2.0$~s). Bold marks the best \flowsim result at each horizon. Time denotes cumulative source-generation time to reach each horizon.}
\label{tab:rollout_sparse}
\centering\footnotesize
\setlength{\tabcolsep}{3.0pt}
\scalebox{0.89}{%
\begin{tabular}{llccccccc}
\toprule
& & \multicolumn{2}{c}{Segmentation} & \multicolumn{2}{c}{Detection} & \multicolumn{2}{c}{Planning} & \multicolumn{1}{c}{Inference}\\
\cmidrule(lr){3-4}\cmidrule(lr){5-6}\cmidrule(lr){7-8}\cmidrule(lr){9-9}
Source & $n$ & Pix.$\uparrow$ & mIoU$\uparrow$ & Box L2$\downarrow$ & Head.$\downarrow$ & ADE$\downarrow$ & FDE$\downarrow$ & Time\,(s)$\downarrow$\\
\midrule
\omnire~\cite{omnire} & -- & 0.77 & 0.42 & 5.78 & 0.84 & 0.61 & 1.60 & 0.45\\
\midrule
\multirow{4}{*}{\shortstack[l]{\flowsim \\(AR,~no assoc.)}}
 & 1 & 0.89 & 0.64 & \textbf{2.36} & 0.48 & 0.44 & 1.17 & 0.06\\
 & 2 & 0.86 & 0.59 & 2.67 & 0.58 & 0.59 & 1.56 & 0.12\\
 & 3 & 0.85 & 0.56 & 2.76 & 0.62 & 0.67 & 1.77 & 0.18\\
 & 4 & 0.85 & 0.55 & 2.76 & 0.62 & 0.71 & 1.89 & 0.24\\
\midrule
\multirow{4}{*}{\shortstack[l]{\flowsim \\(AR,~assoc.)}}
 & 1 & \textbf{0.89} & \textbf{0.65} & \textbf{2.36} & \textbf{0.43} & \textbf{0.38} & \textbf{1.00} & 0.06\\
 & 2 & 0.87 & 0.60 & 2.73 & 0.53 & \textbf{0.49} & \textbf{1.31} & 0.12\\
 & 3 & 0.86 & 0.58 & 2.83 & 0.57 & 0.55 & 1.47 & 0.18\\
 & 4 & 0.85 & 0.56 & 2.84 & 0.60 & 0.58 & 1.56 & 0.23\\
\midrule
\multirow{4}{*}{\shortstack[l]{\flowsim \\(VH,~assoc.)}}
 & 1 & 0.88 & 0.62 & 2.51 & 0.49 & 0.49 & 1.31 & 0.06\\
 & 2 & \textbf{0.88} & \textbf{0.62} & \textbf{2.57} & \textbf{0.50} & 0.50 & 1.35 & \textbf{0.06}\\
 & 3 & \textbf{0.87} & \textbf{0.61} & \textbf{2.60} & \textbf{0.52} & \textbf{0.51} & \textbf{1.38} & \textbf{0.06}\\
 & 4 & \textbf{0.87} & \textbf{0.60} & \textbf{2.66} & \textbf{0.53} & \textbf{0.52} & \textbf{1.40} & \textbf{0.06}\\
\bottomrule
\end{tabular}%
}
\end{table}

\subsection{Logged Conditions and the Symbolic Simulator}\label{sec:scope}
In a complete closed-loop deployment the agent, map, and ego conditions can be supplied by a symbolic scene simulator that updates the scene from the frozen stack's planned trajectory (Fig.~\ref{fig:main}; Sec.~\ref{sec:closedloop_demo}). The quantitative experiments below instead use logged symbolic conditions, which isolate the quality of the latent simulator and provide a real-observation reference for agreement measurement, unavailable for arbitrary counterfactual scenes.

\subsection{Implementation Details}\label{sec:impl}
\flowsim is trained on 4 A100(40GB) GPUs: total batch size 32 for 13 epochs for the dense \uniad instantiation, and 128 for 50 epochs for the sparse \transfuser instantiation. For all experiments on nuScenes, agreement is evaluated over the 150 validation scenes with 3858 common frames. For all experiments on NavSim \texttt{navtest}, evaluation is performed with 147 logged sequences and 6332 common tokens.

\section{EXPERIMENTS}\label{sec:exp}
Our experiments address four questions. \textbf{RQ1}: does \flowsim preserve downstream tasks' agreement as faithfully as image-space simulation sources? \textbf{RQ2}: does explicit cross-frame agent association improve latent transition prediction? \textbf{RQ3}: does variable-horizon training support real-anchored simulation across horizons, and \textbf{RQ4}: how does it differ from autoregressive simulation as error accumulates? We study \uniad with a dense BEV latent on nuScenes and \transfuser with a sparse latent on NavSim; all sources follow Sec.~\ref{sec:eval} for evaluation protocols.

\begin{figure*}[t]
    \centering
    \includegraphics[width=1\linewidth]{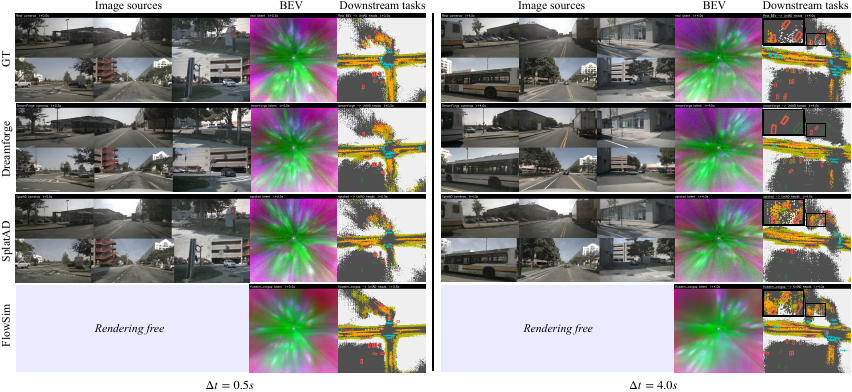}
    \vspace{-6mm}
    \caption{Qualitative comparison of image-space and latent-space simulation on nuScenes, at a short ($\Delta t{=}0.5$s, left) and a long ($\Delta t{=}4.0$s, right) horizon. Each row shows, for a given source, the surround-view image inputs (or GT reference), the resulting BEV feature, and \uniad's downstream detection/lane/drivable-area/planning outputs decoded from it. The green and red boxes are GT annotations~(agent conditions) and predicted results respectively. \dreamforge and \splatad must render an image and re-encode it through \uniad's frontend; \flowsim predicts the BEV latent directly and is rendering-free. At $\Delta t{=}0.5$s all sources remain close to GT, but at $\Delta t{=}4.0$s \dreamforge and \splatad accumulate visible rendering and re-encoding artifacts that degrade the BEV feature and downstream outputs, while \flowsim's latent and decoded outputs stay closer to GT.}
    \label{fig:nuscenes}
    \vspace{-4mm}
\end{figure*}
\begin{figure}
    \centering
    \includegraphics[width=1\linewidth]{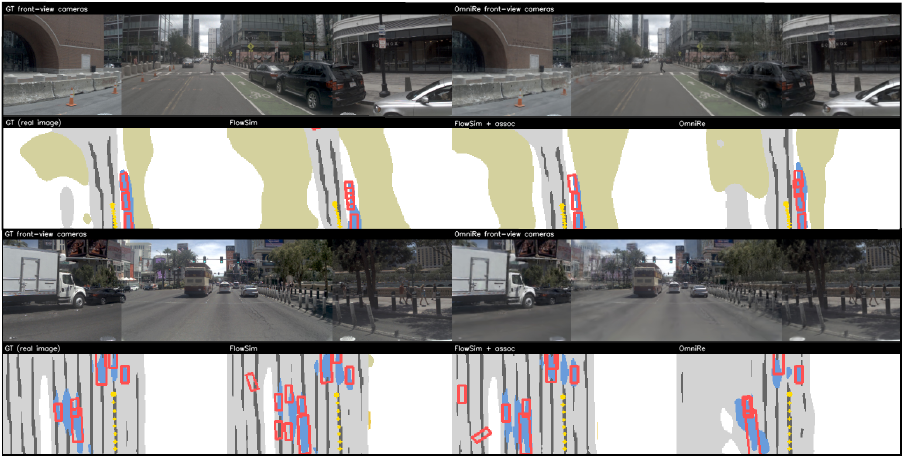}
    \vspace{-6mm}
    \caption{Qualitative single-frame agreement on NavSim \texttt{navtest} (two scenes with $\Delta t=0.5s$). Top row: front-view camera crops used to obtain the source latent for the real GT frame and for \omnire's independently rendered frame. Bottom row: corresponding \transfuser-predicted BEV segmentation, detected agent boxes, and planning results for GT, \flowsim without and with agent association, and \omnire.}
    \label{fig:navsim}
    \vspace{-6mm}
\end{figure}
\vspace{-1mm}
\subsection{Latent-Space vs.\ Image-Space Simulation}\label{sec:openloop}
\paragraph{Dense (nuScenes).}
Table~\ref{tab:agreement_gtprev} compares direct latent prediction with image-space simulation across increasing source-to-target horizons. At $\Delta t{=}0.5$s, when the \uniad receives real history for image-space methods through \uniad's native encoder, the methods exhibit complementary strengths: \dreamforge achieves the lowest planning error ($0.30$/$0.61$m ADE/FDE), \splatad has the highest detection IoU ($0.75$), and \flowsim achieves the lowest detection center error ($1.69$m) while remaining competitive in planning ($0.53$/$0.97$m). The balance shifts at longer horizons, where \uniad must consume generated history from image-space methods. From $1.5$ to $4.0$s, \flowsim with variable-horizon~(VH) simulation and agent association~(assoc.) outperforms \dreamforge on every reported metric and surpasses generated-history \splatad on nearly all metrics. At $4.0$s, for example, \flowsim improves detection IoU from $0.36$ to $0.55$ and reduces planning ADE/FDE from $0.99$/$2.07$m to $0.66$/$1.23$m relative to \dreamforge. \flowsim is also consistently the cheapest source: its single latent-transition step takes only $0.04$s and $3.0$~GB of VRAM at every horizon, versus $0.13$s/$4.5$~GB for \splatad and $92.3$--$726.0$s/up to $40.0$~GB for \dreamforge.

Fig.~\ref{fig:nuscenes} qualitatively explains this difference. \dreamforge must generate an image sequence autoregressively, so generation errors accumulate in its history before being re-encoded by \uniad. \splatad avoids this temporal extrapolation but still incurs a rendering-to-perception domain gap when its rendered observations are re-encoded. In contrast, \flowsim predicts the BEV latent directly from a common real anchor and the target symbolic conditions, avoiding image-space simulation. Although its predicted BEV features appear smoother in the visualization, the resulting downstream tasks' outputs remain close to those predicted by real observations. Thus, preserving all observation-level feature variation is not necessary for preserving the downstream behavior relevant to simulation-based evaluation.

\paragraph{Sparse (NavSim).}
The same conclusion holds for the sparse latent of \transfuser. In the single-frame NavSim setting, \flowsim with agent association outperforms the image-space \omnire baseline on every reported metric (Table~\ref{tab:rollout_sparse}), improving segmentation mIoU from $0.42$ to $0.65$ and reducing planning ADE from $0.61$ to $0.38$m, while also simulating each step in $0.06$s versus \omnire's $0.45$s. Fig.~\ref{fig:navsim} shows the corresponding qualitative behavior: \omnire's rendered-then-re-encoded observations blur lane boundaries and shift agent boxes, whereas \flowsim preserves both directly in latent space. This result across dense and sparse representations supports the central premise that decision-consistent latent simulation can be a more faithful and efficient alternative to image-space simulation for preserving downstream tasks' behavior in simulation.
\vspace{-1mm}
\subsection{Variable-Horizon Simulation and Agent Association}\label{sec:temporalbehav}
\emph{Real-anchored variable-horizon simulation avoids autoregressive drift (RQ3--RQ4).}
We compare two ways of reaching a future horizon. In autoregressive (AR) simulation, each generated latent is fed back as the input to the next prediction, so errors introduced at one step affect all subsequent predictions. In variable-horizon (VH) simulation, each target latent is instead predicted directly from a logged real anchor and the symbolic conditions at the queried horizon. Thus, VH isolates horizon-dependent simulation fidelity from the additional error introduced by recursive sim-to-sim rollouts.

Across both latent representations, this distinction becomes larger as the horizon increases. On dense \uniad latents (Table~\ref{tab:multihorizon_dense}), AR simulation yields planning FDEs of $1.32$, $1.42$, and $1.45$m at $1.5$, $2.5$, and $4.0$s, respectively, whereas VH prediction remains lower at $1.14$, $1.13$, and $1.23$m. The sparse \transfuser results show the same trade-off more clearly (Table~\ref{tab:rollout_sparse}). AR is stronger for the immediate transition, achieving $0.38$m planning ADE at $n{=}1$, compared with $0.49$m for VH. However, AR's error accumulates from $0.38$ to $0.58$m by $n{=}4$, while VH remains nearly constant from $0.49$ to $0.52$m and becomes preferable from the later simulation steps onward. These results show that VH training and simulation trade some short-horizon fidelity for substantially reduced long-horizon drift.
\begin{figure*}[t]
    \centering
    \includegraphics[width=1\linewidth]{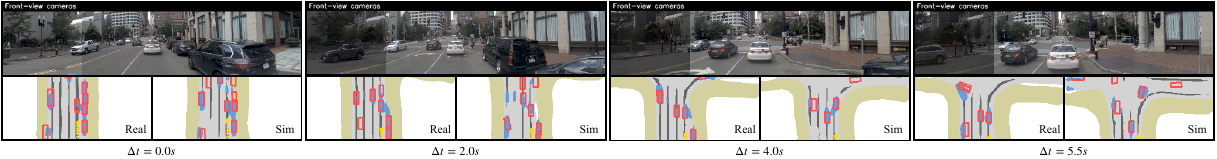}
    \vspace{-6mm}
    \caption{Counterfactual closed-loop rollout illustrated in Sec.~\ref{sec:closedloop_demo}. We show four sampled steps of the $5.5$s rollout; each panel shows the front-view cameras together with the BEV segmentation decoded from the logged real latent (\emph{Real}) and from \flowsim's closed-loop simulated latent (\emph{Sim}).}
    \label{fig:closed_loop}
\end{figure*}
\emph{Agent association improves dynamic transition prediction and rollout stability (RQ2).}
We compare AR simulation with and without associated-agent conditioning. On dense \uniad latents, the two variants are comparable at short horizons, but their behavior diverges sharply under repeated feedback. At $4.0$s, removing association increases planning ADE/FDE from $0.77$/$1.45$m to $1.97$/$3.76$m and increases detection center error from $2.19$ to $4.80$m (Table~\ref{tab:multihorizon_dense}). The sparse \transfuser results further show that association improves the transition itself, not only its long-horizon accumulation: at $n{=}1$, it reduces planning ADE from $0.44$ to $0.38$m and heading error from $0.48$ to $0.43$. Its benefit persists throughout rollout, reducing planning ADE from $0.71$ to $0.58$m at $n{=}4$ (Table~\ref{tab:rollout_sparse}). Explicit source--target correspondence therefore allows the simulator to model agent motion directly rather than infer it from independent frame-wise scene descriptions, improving both simulation agreement and stability.

\begin{table}[t]
\caption{\flowsim rollout ablation with \uniad at four horizons: two \emph{autoregressive} variants (with/without associated agents) that re-feed each generated latent, and the \emph{variable-horizon} variant reproduced from Table~\ref{tab:agreement_gtprev}. Bold marks the best per horizon.}
\vspace{-1mm}
\label{tab:multihorizon_dense}
\centering\footnotesize
\setlength{\tabcolsep}{3.0pt}
\scalebox{0.9}{%
\begin{tabular}{llcccccc}
\toprule
& & \multicolumn{2}{c}{Detection} & \multicolumn{2}{c}{Map} & \multicolumn{2}{c}{Planning}\\
\cmidrule(lr){3-4}\cmidrule(lr){5-6}\cmidrule(lr){7-8}
Protocol & $\Delta t$ & IoU$\uparrow$ & cADE$\downarrow$ & Lane$\uparrow$ & Driv.$\uparrow$ & ADE$\downarrow$ & FDE$\downarrow$\\
\midrule
\multirow{4}{*}{\shortstack[l]{\flowsim \\(AR, no assoc.)}}
 & $0.5$~s & \textbf{0.69} & \textbf{1.41} & \textbf{0.49} & \textbf{0.86} & 0.61 & 1.19 \\
 & $1.5$~s & \textbf{0.58} & 2.13 & 0.36 & 0.77 & 1.25 & 2.45 \\
 & $2.5$~s & 0.54 & 2.77 & 0.28 & 0.70 & 1.60 & 3.10 \\
 & $4.0$~s & 0.46 & 4.80 & 0.19 & 0.61 & 1.97 & 3.76 \\
\midrule
\multirow{4}{*}{\shortstack[l]{\flowsim \\(AR, assoc.)}}
 & $0.5$~s & 0.64 & 1.76 & \textbf{0.49} & \textbf{0.86} & \textbf{0.48} & \textbf{0.92} \\
 & $1.5$~s & 0.56 & 2.02 & 0.40 & 0.81 & 0.67 & 1.32 \\
 & $2.5$~s & 0.55 & \textbf{2.12} & 0.37 & 0.79 & 0.74 & 1.42 \\
 & $4.0$~s & 0.54 & 2.19 & 0.34 & 0.77 & 0.77 & 1.45 \\
 \midrule
\multirow{4}{*}{\shortstack[l]{\flowsim \\(VH, assoc.)}}
 & $0.5$~s & 0.68 & 1.69 & 0.48 & \textbf{0.86} & 0.53 & 0.97 \\
 & $1.5$~s & \textbf{0.58} & \textbf{1.99} & \textbf{0.43} & \textbf{0.82} & \textbf{0.61} & \textbf{1.14} \\
 & $2.5$~s & \textbf{0.56} & \textbf{2.12} & \textbf{0.40} & \textbf{0.81} & \textbf{0.60} & \textbf{1.13} \\
 & $4.0$~s & \textbf{0.55} & \textbf{2.18} & \textbf{0.38} & \textbf{0.79} & \textbf{0.66} & \textbf{1.23} \\
\bottomrule
\end{tabular}
}
\vspace{-3mm}
\end{table}

\subsection{Toward Closed-Loop Evaluation}\label{sec:closedloop_demo}
While Secs.~\ref{sec:openloop} and \ref{sec:temporalbehav} use logged symbolic conditions for real-referenced agreement, \flowsim's conditioning interface is agnostic to their source and can instead be driven by a reactive symbolic simulator. Fig.~\ref{fig:closed_loop} demonstrates this using VH simulation with \transfuser on a counterfactual scenario: a leading vehicle that originally stops at a red light is edited to keep driving. At each step, \flowsim uses the geometrically nearest logged real frame as the source latent anchor, while the agent, map, and ego conditions are instead updated online by the symbolic simulator from \flowsim's own previous planned trajectory rather than the log. Over the $5.5$s rollout, \flowsim continues to produce plausible, coherent outputs that diverge from the logged behavior as the counterfactual takes effect, illustrating how VH simulation supports closed-loop deployment by allowing the scene conditions to be updated between simulation steps.

\section{CONCLUSION AND LIMITATIONS}
We presented \textsc{FlowSim}, a rendering-free latent simulation paradigm for E2E driving systems that predicts future latents directly from a previous latent and symbolic scene conditions. To support stable latent transitions and long-horizon simulation, we introduced explicit agent association and real-anchored variable-horizon simulation. Using a source-to-reference agreement protocol, we showed that \flowsim can match or exceed image-space simulation baselines in downstream agreement while running significantly faster.

A limitation of this design is that \flowsim is tied to the latent space of a specific frozen perception encoder, requiring a corresponding simulator to be trained for each target E2E stack. 
While \flowsim does not re-execute the perception encoder during simulation, its supervision is derived from the encoder's latent representations on real sensor observations. The simulator is therefore trained to reproduce the latent that the encoder would have produced under the target scene conditions, reflecting the encoder's behavior for simulation.

Taken together, \flowsim serves as a latent-space interface for closed-loop evaluation without image rendering, and we hope it provides a step toward more efficient and scalable evaluation of E2E driving systems.

\end{document}